%% file: root.tex
\documentclass[letterpaper, 10 pt, conference]{ieeeconf}

\IEEEoverridecommandlockouts                              

\usepackage{multirow}
\usepackage{graphicx} 

\usepackage{url}
\usepackage{xcolor}
\usepackage{algorithm}
\usepackage[noend]{algpseudocode} 
\usepackage{amssymb,amsmath,comment,cite}
\usepackage{booktabs}
\usepackage{tabularx}
\usepackage{threeparttable}
\usepackage{array}
\usepackage{makecell}

\usepackage[bookmarks=true]{hyperref}

\usepackage{amsthm} 

\newtheorem{assumption}{Assumption}
\newtheorem{proposition}{Proposition}

\usepackage{ifthen}
\newboolean{shortver}
\setboolean{shortver}{true}

\title{\bf S-Rolling: Steerable Rolling Gaits for Twistable Snake Robots in Low-Clearance Environments}

\author{Qingling Fan$^{1}$, Zhongqiang Ren$^{1\dagger}$ 
\thanks{$^{1}$Global College, Shanghai Jiao Tong University, China (\texttt{\{qingling.fan, zhongqiang.ren\}@sjtu.edu.cn}).}
\thanks{$^\dagger$ Corresponding author}
}

\usepackage{xspace}

\usepackage{etoolbox}
\makeatletter
\patchcmd{\@makecaption}
  {\scshape}
  {}
  {}
  {}
\patchcmd{\@makecaption}
{\\}
{.\ }
{}
{}
\makeatother

\begin{document}

\maketitle

\graphicspath{{./figures/}}

\newcommand{\abbrTSnake}{T-Snake\xspace}

\begin{abstract}
    \input{abstract}
\end{abstract}

\graphicspath{{./figure/}}
 
\section{Introduction}
\input{intro2}

\section{Related Works}\label{sec:related_works}
\input{related}

\section{Method}\label{sec:method}
\input{method}

\section{Experimental Results}\label{sec:experiments}
\input{experiments}

\section{Conclusion and Future Works}\label{sec:conclude}
\input{conclude}

\section*{Acknowledgments}

OpenAI Codex assisted with plotting code for Fig.~\ref{fig:dorsal_ref_curve}, and with grammar checking and wording revisions.
The authors take responsibility for the final content.

\bibliographystyle{IEEEtran}
\bibliography{references}

\end{document}

%% file: abstract.tex
Snake robots can traverse complex environments with limbless
locomotion, but maneuvering in low-clearance environments remains challenging because a gait must maintain a small swept height while also providing planar steering. 
Existing rolling gaits preserve a low swept height body configuration but generally provide limited steering, whereas steering gaits often introduce vertical undulation and require additional clearance. 
With that in mind, this paper develops S-rolling, a steerable rolling gait that combines low swept height locomotion with planar maneuverability for twistable snake robots.
S-rolling is designed using an S-shaped planar backbone curve, whose head and tail segments roll with independently controlled angular velocities, producing different motion modes. 
We derive its joint space gait equation, in which a central twist joint realizes the relative rotation between the two segments. 
We also develop a quasi-static kinematic model relating the segment rolling speeds to the linear and angular velocities of the robot. 
Real-robot experiments demonstrate the various motion modes, traversal through a $10$ cm clearance test track, and maneuvering tasks beneath a large workpiece and a car chassis. 
In these low-clearance scenarios, S-rolling successfully performs approximately $90^\circ$ and $180^\circ$ heading changes, whereas the baseline rolling arc gait cannot change the robot heading. 

%% file: intro2.tex
\begin{figure}[tb]
    \centering
    \includegraphics[width=1.0\linewidth]{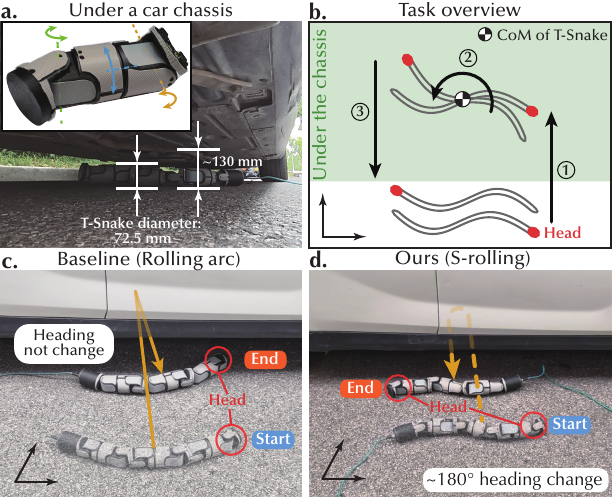}
    \caption{The \abbrTSnake uses S-rolling to conduct the tests under a car chassis. 
    (a) The joint arrangement of a \abbrTSnake module, and the bottom view of the \abbrTSnake traversing under the chassis.
    (b) Task sequence: 1) entering beneath the vehicle; 2) turning $180^\circ$ in place to reorient its head; and 3) exiting from beneath the vehicle.
    (d) The proposed S-rolling gait realizes $180^\circ$ heading change, while (c) the baseline gait, rolling arc, cannot perform.}
    \label{fig:head_figure}
\end{figure}

The paper is motivated by robot navigation in Low-Clearance Environments (LCE), by which we refer to expansive horizontal surfaces that have restricted vertical space (i.e., a ceiling) above them.
Examples of LCEs include the space beneath a vehicle chassis (Fig.~\ref{fig:head_figure}), and navigating robots in LCEs naturally arises in applications such as exploration or inspection of vehicles or large industrial parts.
Navigating LCEs is quite challenging: on the one hand, the limited vertical space requires the robot to have a small cross section to be able to enter LCEs.
On the other hand, the robot should be easily maneuverable, such as moving back and forth and changing directions in LCEs.

Snake robots, as a class of hyper-redundant mechanisms, enjoy a small cross section and can traverse complex environments such as pipes, narrow passages, and compact spaces with limbless locomotion~\cite{pettersen2017snake,richter2022arcsnakea,vaquero2024eels}.
However, due to the limited vertical space of LCEs, a snake robot must maintain a small swept height to avoid collision with the ceiling, which makes many existing gaits, cyclic motion in the shape space, for snake robots inapplicable:
To name a few, concertina gaits rely on lateral anchoring and may require sufficient contact with the surrounding walls~\cite{transeth2009survey,koley2025adaptivea}, which may not be available in LCEs.
Sidewinding can generate steerable motion, but its vertical undulation increases the swept height and may collide with the ceiling in LCEs~\cite{tanev2005automated,gong2013extended}.

Possible gaits for snake robots to locomote in LCEs include rolling and rectilinear gaits, which maintain a small swept height.
However, these gaits generally provide limited steering capability~\cite{zhen2015modelinga,tang2015studya,chen2004studies,hopkins2014design}: the robot can only move back and forth along a straight line without changing direction, which limits its maneuverability in LCEs.

This paper is interested in enabling the snake robot to achieve steerable motion in LCEs, without losing existing gaits for the snake robot to traverse complex terrains.
To generate steerable motion in LCEs, this paper considers a so-called twistable snake robot (\abbrTSnake)~\cite{fan2026motion} and develops a new \emph{S-rolling} gait for \abbrTSnake.
\abbrTSnake is a mechanism that can not only bend its backbone as the popular modular snake robot does~\cite{rollinson2014design}, which allows \abbrTSnake to achieve any gait that regular snake robot has, but also twist about its backbone (Fig.~\ref{fig:head_figure}(a)), which provides new possibilities for locomotion.
Based on such \abbrTSnake mechanism, we take the view that low swept height and planar maneuverability can be achieved simultaneously by coordinating the rolling motions of different body segments, without introducing additional vertical undulation.
With that in mind, the proposed S-rolling gait leverages an S-shaped backbone curve, which consists of two tangential circular arcs with opposite signed curvatures.
The two segments of \abbrTSnake can roll with independently controlled angular velocities, which thus enables steerable motion in LCEs.

Designing and controlling such new gaits for \abbrTSnake faces the fundamental challenge of coordinating the many degree of freedom (DoFs) of a snake robot to generate effective motion with real-time computation.
To address the challenge, we begin our design of S-rolling using the backbone curve approach~\cite{chirikjian1994modal,yamada2006studya}, which provides an intuitive geometric method to model and design the gait of the snake robot.
Then, we further derive the gait equations, closed-form shape space motion, of S-rolling based on the designed backbone curve, which allows light-weight and fast computation.
With the help of the S-rolling gait, we further derive a quasi-static kinematic model that relates the two segment rolling speeds to the linear and angular velocities of the robot, which allows locomotive reduction of a complicated snake robot to a simple differential drive car, for easy maneuver in practice.

To summarize, the technical contributions of this paper include: (i) the development of S-rolling gaits, which include both the design of the backbone curve and the further derivation of gait equations, and (ii) the quasi-static locomotive reduction of \abbrTSnake as a differential-drive car based on S-rolling that enables easy maneuver in LCEs.

For verification, we conduct real-robot experiments with \abbrTSnake hardware.
We first demonstrate forward and backward translation, clockwise and counterclockwise in-place turning. 
We then evaluate the gait on a test track containing $10$ cm clearance passages.
Finally, we demonstrate two practical tasks beneath a large workpiece and a car chassis, with clearance heights of $15$ cm and $13$ cm, respectively. 
Compared with the rolling arc gait~\cite{zhen2015modelinga}, which cannot change the robot heading in these tasks (Fig.~\ref{fig:head_figure}(c)), S-rolling successfully performs approximately $90^\circ$ and $180^\circ$ heading changes while maintaining low-clearance locomotion.

%% file: related.tex
\subsection{Gait Design Methods}\label{subsec:gait_designs}

Designing snake-robot gaits is difficult because many redundant DoFs couple shape-space motion to workspace displacement. 
Central pattern generators~\cite{yu1999sensory,ijspeert2007swimming,wang2017cpginspired} and reinforcement learning~\cite{liu2023reinforcement,jiang2024hierarchical,bing2022simulation} improve adaptability, but often provide limited geometric interpretability. 
Parameterized gait equations simplify redundant-joint control for common gaits~\cite{tesch2009parameterizedb,hatton2010sidewinding,zhen2015modelinga}.
This approach offers low computational cost and high real-time performance.
Backbone curve methods~\cite{chirikjian1994modal,yamada2006studya} encode body shape and orientation geometrically, enabling diverse rolling and climbing gaits~\cite{zhen2015modelinga,takemori2023adaptivea,tanaka2018developmenta}. 
We use this advantage to design S-rolling and derive its joint-space equation.


\subsection{Rolling and Low-Clearance Locomotion}

Rolling-based gaits are developed for transport, terrain traversal, pipe climbing, and manipulation~\cite{miki2026snake,zhen2015modelinga,takemori2023adaptivea,elsayed2022mobilea}. 
These existing rolling gaits typically prescribe a fixed rolling direction.
In particular, the rolling arc gait~\cite{zhen2015modelinga}, considered as the baseline in this paper, provides low swept height locomotion, but does not provide any steering capability.


The proposed gait targets low-clearance environments, which constrain the gravity direction but not the horizontal plane. 
Let $h_z$ denote the height of a narrow gap, and $A_{cd}$ denote the cross-sectional diameter of the snake robot.
We define an LCE by $1\leq h_z/A_{cd}<2$, analogous to the restricted-maneuverability criterion in~\cite{murphy2017disaster,xiao2018review}.

Concertina gaits~\cite{transeth2009survey,koley2025adaptivea} need lateral anchoring, while ceiling-floor anchoring consumes the available height. 
Planar rolling and rectilinear gaits keep swept height low.
But steering of these gaits requires additional vertical body waves or lifting~\cite{zhen2015modelinga,tang2015studya,chen2004studies,hopkins2014design}.
Sidewinding steers effectively but increases swept height~\cite{tanev2005automated,gong2013extended}.

%% file: method.tex
\begin{figure}[tb]
    \centering
    \includegraphics[width=1.0\linewidth]{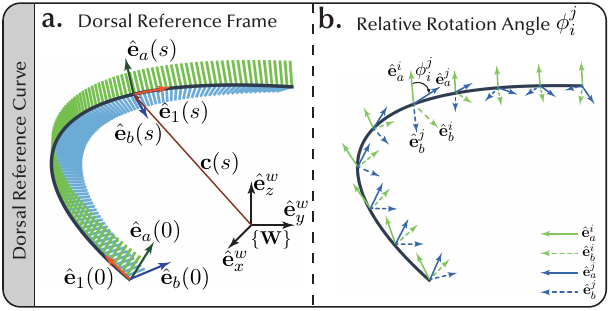}
    \caption{Overview of the dorsal reference curve. (a) The dorsal reference frames along the curve $\mathbf{c}(s)$. (b) The relative rotation angle $\phi_i^j$ from frame $\mathbf{D}^i$ to frame $\mathbf{D}^j$.}
    \label{fig:dorsal_ref_curve}
\end{figure}

In this section, we first design the S-rolling gait using the backbone curve method (Sec.~\ref{subsec:backbone_curves}). 
We then describe the discrete mechanical structure of \abbrTSnake and its joint arrangement (Sec.~\ref{subsec:mechanism}). 
Based on the resulting S-rolling backbone curve and the kinematic structure of \abbrTSnake, we use the integration method to fit the curve, deriving a low-dimensional representation of the gait, referred to as the gait equations (Sec.~\ref{subsec:gait_eqn}). 
We subsequently develop a kinematic model of S-rolling (Sec.~\ref{subsec:kinematic_model}). 

\subsection{Design of Backbone Curves}\label{subsec:backbone_curves}

The geometric shape and orientation of the snake robot body can be designed by the backbone curve method~\cite{chirikjian1994modal,yamada2006studya}.
The commonly used Frenet-Serret frames describe the shape of the curve, yet lack the ability to define the rotation along the curve~\cite{yamada2006studya}.
Instead, the dorsal reference frame~\cite{yamada2006studya} can describe not only the geometric shape, but also the rotation along the curve, which is needed in this paper.

\subsubsection{Dorsal Reference Curve}

Let $\{\mathbf{W} \}=[\hat{\mathbf{e}}_x^w,\hat{\mathbf{e}}_y^w,\hat{\mathbf{e}}_z^w]^\top \in  \mathbb{R}^{3\times3}$ denote the world frame of a 3D workspace, and let $\mathbf{c}(s)=[x(s),y(s),z(s)] \in \mathbb{R}^{1\times3}$ denote a spatial curve represented in this world frame, which is parameterized by its arc-length $s\in[0,L]$.
The dorsal reference frames $\mathbf{D}=[\hat{\mathbf{e}}_1,\hat{\mathbf{e}}_a,\hat{\mathbf{e}}_b]^\top \in \mathbb{R}^{3\times3}$ are defined along the curve $\mathbf{c}(s)$ (Fig.~\ref{fig:dorsal_ref_curve}(a)) with three orthonormal vectors, where $\hat{\mathbf{e}}_1$ is the unit tangent vector along the curve $\mathbf{c}(s)$, and $\hat{\mathbf{e}}_a$, $\hat{\mathbf{e}}_b$ are the unit normal and binormal vectors along the curve, respectively.
The dorsal reference frames along the curve satisfy the following differential equations~\cite{yamada2006studya}:
\begin{subequations}\label{eqn:shape_equation}
\begin{align}
    \frac{{\rm d} \mathbf{c}}{{\rm d} s} &= \hat{\mathbf{e}}_1 \label{eqn:tangent} \\
    \frac{{\rm d} \mathbf{D}}{{\rm d} s} &= \begin{bmatrix}
    0 & \kappa_b(s) & -\kappa_a(s) \\
    -\kappa_b(s) & 0 & \tau(s) \\
    \kappa_a(s) & -\tau(s) & 0
    \end{bmatrix} \mathbf{D} \label{eqn:shape_eqn_matrix}
\end{align}
\end{subequations}

Equation~\eqref{eqn:shape_eqn_matrix} is called the \textit{shape equation} of the dorsal reference curve, where $\kappa_a$, $\kappa_b$ and $\tau$ are termed \textit{shape functions}.
Here, $\kappa_a$ and $\kappa_b$ represent the curvatures along two axes perpendicular to the tangent axis of the curve, while $\tau$ represents the torsion along the curve.

\subsubsection{Rolling Description}

When $\tau=0$ for the entire curve, the corresponding dorsal reference curve is called the \textit{bellows model}, which describes the case that the snake robot is able to bend but unable to twist.
We call the dorsal reference frame on the bellows model the \textit{bellows frame}.
Since a bellows model does not allow torsion along the curve, when any of bellows frames along the curve rotate about the curve's tangent vector, all other frames must rotate correspondingly so as to ensure torsion-free backbones.
In other words, this relative rotation angle $\phi$ is spatially invariant along the entire curve, and is independent of the arc-length $s$, i.e., $\phi(s)=\phi_0$.

As shown in Fig.~\ref{fig:dorsal_ref_curve}(b), consider two bellows frames attached on the same backbone curve $\mathbf{c}(s)$ at the same point: $\mathbf{D^i}=[\hat{\mathbf{e}}_1^i,\hat{\mathbf{e}}_a^i,\hat{\mathbf{e}}_b^i]^\top$ and $\mathbf{D^j}=[\hat{\mathbf{e}}_1^j,\hat{\mathbf{e}}_a^j,\hat{\mathbf{e}}_b^j]^\top$. 
Specifically, the tangent vectors of frames $\mathbf{D^i}$ and $\mathbf{D^j}$ are the same for any point along the curve $s\in[0,L]$ (i.e., $\hat{\mathbf{e}}_1^i=\hat{\mathbf{e}}_1^j$).
We use notation $\phi_i^j(s)$ to denote the relative rotation angle from frame $\mathbf{D^i}$ to frame $\mathbf{D^j}$.
Then, the shape functions for these two frames satisfy the following equations~\cite{yamada2006studya}:
\begin{subequations}\label{eqn:shape_eqns_relation}
    \begin{align}
        \begin{bmatrix}
            \kappa_a^j \\
            \kappa_b^j
        \end{bmatrix} &= 
        \begin{bmatrix}
            \cos{\phi_i^j} & \sin{\phi_i^j} \\
            -\sin{\phi_i^j} & \cos{\phi_i^j}
        \end{bmatrix} 
        \begin{bmatrix}
            \kappa_a^i \\
            \kappa_b^i
        \end{bmatrix} \label{eqn:rotation_angle} \\
        \tau^j &= \tau^i = 0, \label{eqn:torsion}
    \end{align}
\end{subequations}
which are leveraged to design S-rolling gait later.
By designing $\phi_i^j(t)$ as a time-varying function, the resulting curve can generate lateral rolling motions of snake robots~\cite{yamada2006studya,zhen2015modelinga}.

\begin{figure}[tb]
    \centering
    \includegraphics[width=1.0\linewidth]{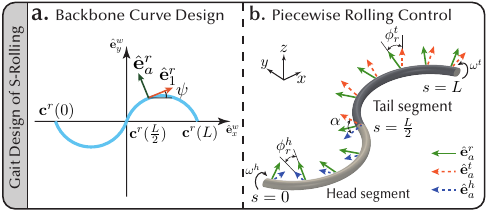}
    \caption{Develop S-rolling using backbone curve method. (a) The backbone curve consists of two tangential circular arcs. A bellows frame $\mathbf{D}^r$ is defined along the curve. We add geometric constraints to determine the unique configuration of the curve. (b) The curve is divided into two segments, and the rolling of each segment is controlled by a different rotation angle. After rotating $\hat{\mathbf e}_a^r$, it forms $\hat{\mathbf e}_a^h$ in frame $\mathbf D^h$ and $\hat{\mathbf e}_a^t$ in frame $\mathbf D^t$. }
    \label{fig:backbone_curve_design}
\end{figure}

\subsubsection{Backbone Curve of S-Rolling}

We design the backbone curve for S-rolling, denoted by $\mathbf{c}^r(s):s\in[0,L]\to \mathbb{R}^{1\times3}$, to lie in the $XoY$ plane extended by $\hat{\mathbf e}_x^w$ and $\hat{\mathbf e}_y^w$ of the world frame.
As shown in Fig.~\ref{fig:backbone_curve_design}(a), the curve consists of two tangential circular arcs of equal length and equal curvature magnitude but opposite signed curvatures.
Geometrically, the resulting curve resembles the letter ``s", and hence the name \textit{S-rolling}.
The resulting closed-form parametric representation, given in Eq.~\eqref{eqn:curve_of_s_rolling}, serves as the basis for subsequent kinematic analysis.

We now derive this expression and first construct a bellows frame $\mathbf D^r$ along $\mathbf{c}^r(s)$.
Since $\mathbf{c}^r$ lies in the $XoY$ plane, its unit tangent vector $\hat{\mathbf e}_1^r$ also lies in this plane.
We further choose the normal vector $\hat{\mathbf e}_a^r$ to lie in the same plane.
According to the right-hand rule, the binormal vector satisfies $\hat{\mathbf e}_b^r
=
\hat{\mathbf e}_1^r\times\hat{\mathbf e}_a^r
\parallel \hat{\mathbf e}_z^w.$
Then, the shape functions associated with this frame satisfy:
\begin{equation}\label{eqn:kappa_a_and_tau_0}
    \kappa_a^r(s)=0,\qquad \tau^r(s)=0,
\end{equation}
while $\kappa_b^r(s)$ is a piecewise constant function with respect to the arc-length parameter $s$:
\begin{equation}\label{eqn:piecewise_kappa_b}
    \kappa_b^r(s)=
    \begin{cases}
        \displaystyle
        \kappa_0, & 0\leq s< \frac{L}{2},\\[1mm]
        \displaystyle
        -\kappa_0, & \frac{L}{2}<s\leq L.
    \end{cases}
\end{equation}

The shape equations derived so far determine the intrinsic geometry of the curve but not its absolute position and orientation in the world frame $\{\mathbf{W} \}$.
 As shown in Fig.~\ref{fig:backbone_curve_design}(a), to obtain a unique representation of $\mathbf c^r$ in the world frame, we place the arc-length midpoint $\mathbf c^r(L/2)$ at the origin of the world frame.
We further require the initial endpoint $\mathbf c^r(0)$ to lie on the negative $x$-axis defined by $\hat{\mathbf e}_x^w$, and the terminal endpoint $\mathbf c^r(L)$ to lie on the positive $x$-axis.
These conditions eliminate the arbitrary translation and rotation of the curve when being represented in the world frame, thereby yielding a unique parametric representation of $\mathbf c^r(s)$.

To derive the closed-form parametric expression of the backbone curve, we then define the signed tangent angle $\psi(s)$ as the angle measured from the $+x$-axis to the tangent vector $\hat{\mathbf e}_1^r(s)$ (see Fig.~\ref{fig:backbone_curve_design}(a)).
Using this tangent angle, we decompose the tangent and normal vectors of the frame $\mathbf D^r$ into their $x$- and $y$-components:
\begin{equation}\label{eqn:psi_to_vectors}
        \hat{\mathbf e}_1^r
        =\cos\psi\,\hat{\mathbf e}_x^w
        +\sin\psi\,\hat{\mathbf e}_y^w,
        \qquad
        \hat{\mathbf e}_a^r
        =-\sin\psi\,\hat{\mathbf e}_x^w
        +\cos\psi\,\hat{\mathbf e}_y^w.
\end{equation}
Substituting these expressions into the shape equation, i.e., Eq.~\eqref{eqn:shape_eqn_matrix}, yields 
\begin{equation}\label{eqn:d_tangent_angle}
    \frac{{\rm d} \psi}{{\rm d} s}=\kappa_b^r(s).
\end{equation}
From the geometric requirements that both endpoints of the curve lie on the $x$-axis, we obtain the conditions $\psi\left({L}/{4}\right) =\psi\left({3L}/{4}\right)=0$.
Integrating Eq.~\eqref{eqn:d_tangent_angle} subject to these conditions gives the following expression for the tangent angle:
\begin{equation}
    \label{eqn:piecewise_tangent_angle}
    \psi(s)=
    \begin{cases}
        \displaystyle
        \kappa_0\left(s-\frac{L}{4}\right),
        &0\leq s\leq \dfrac L2,\\[2mm]
        \displaystyle
        \kappa_0\left(\frac{3L}{4}-s\right),
        &\dfrac L2\leq s\leq L.
    \end{cases}
\end{equation}
Substituting Eq.~\eqref{eqn:piecewise_tangent_angle} back into Eq.~\eqref{eqn:psi_to_vectors} gives the tangent vector $\hat{\mathbf e}_1^r(s)$. 
Then, according to Eq.~\eqref{eqn:tangent}, the backbone curve is obtained by integrating the tangent vector with respect to the arc-length parameter $s$. 
Imposing the midpoint condition $\mathbf c^r(L/2)=\mathbf 0$ finally yields the parametric expression for the backbone curve $\mathbf c^r(s)$:
\begin{equation}
    \label{eqn:curve_of_s_rolling}
    \begin{aligned}
        \mathbf c^r(s)
    &=
    \frac{\sigma(s)}{\kappa_0}
    \left[
        \sin\psi(s)-\sin\psi_m,\,
        \cos\psi_m-\cos\psi(s),\,
        0
    \right], \\
    \sigma(s)
    &=
    \begin{cases}
        1,  & 0\leq s\leq \frac {L}{2},\\
        -1, & \frac {L}{2} <s\leq L,
    \end{cases}
    \qquad
    \psi_m=\frac{\kappa_0L}{4}.
    \end{aligned}
\end{equation}
Here, $\psi_m$ is defined for simplicity and $\sigma(s)$ distinguishes the two circular arcs.
The closed-form parametric expression of the S-rolling backbone curve, $\mathbf c^r(s)$, and the normal vector $\hat{\mathbf e}_a^r$ of the frame $\mathbf D^r$ will be used to derive the kinematic model in Sec.~\ref{subsec:kinematic_model}.

\subsubsection{Piecewise Rolling Control and Twist Angle}

Recall that, for a bellows frame, controlling its relative rotation about the tangent vector enables the snake robot body to perform lateral rolling motions~\cite{yamada2006studya}.
The two circular arcs of the S-rolling backbone curve $\mathbf c^r(s)$ naturally divide the robot body into two parts.
We define $s\in[0,L/2]$ as the \textit{head segment} and $s\in(L/2,L]$ as the \textit{tail segment}.
For the head segment, the reference bellows frame $\mathbf D^r$ is rotated about its tangent vector through a relative rotation angle $\phi_r^h(t)$, resulting in a new bellows frame denoted by $\mathbf D^h(s,t),s\in [0,L/2]$ (Fig.~\ref{fig:backbone_curve_design}(b)). 
Similarly, for the tail segment, $\mathbf D^r$ is rotated about its tangent vector through an angle $\phi_{r}^t(t)$, resulting in the bellows frame $\mathbf D^t(s,t),s\in(L/2,L]$ (Fig.~\ref{fig:backbone_curve_design}(b)).

We further define the time derivatives of the two rotation angles as the rolling speeds:
\begin{equation}
    \label{eqn:rolling_speeds}
    \omega^h =\dot{\phi}_r^h, \quad \omega^t = \dot{\phi}_r^t,
\end{equation}
where $\omega^h$ and $\omega^t$ are the \textit{head rolling speed} and the \textit{tail rolling speed}, respectively.
We can compute the shape function of new frames $\mathbf D^h$ and $\mathbf D^t$ using Eq.~\eqref{eqn:rotation_angle}.
To obtain a concise expression, we consider that both rotation angles vary at constant rates:
\begin{subequations}\label{eqn:new_shape_functions}
\begin{align}
    \kappa_a^h(t) &=\kappa_0 \sin{\left( \omega^h t \right)}, \quad \kappa_b^h(t) =\kappa_0 \cos{\left( \omega^h t \right)} \\ 
    \kappa_a^t(t) &=-\kappa_0 \sin{\left( \omega^t t \right)}, \quad \kappa_b^t(t) =-\kappa_0 \cos{\left( \omega^t t \right)}.
\end{align}
\end{subequations}
These functions will be used to derive the gait equation of S-rolling further in Sec.~\ref{subsec:gait_eqn}.

By independently controlling two rotation angles $\phi_r^h$ and $\phi_r^t$ and their rates of change, the head and tail segments of the snake robot can perform rolling motions with different directions and speeds, thereby generating different locomotion modes, as will be further analyzed in Sec.~\ref{subsec:kinematic_model}.
It should be noted that when the two rotation angles are unequal, the frames $\mathbf D^h$ and $\mathbf D^t$ are no longer continuous across the segment junction.
In particular, a finite jump in rotation about the tangent vector occurs at $s=L/2$ (Fig.~\ref{fig:backbone_curve_design}(b)).
We define this relative rotation as the \textit{twist angle}:
\begin{equation}
    \label{eqn:twi_angle}
    \alpha(t)
        =
        \phi_{r}^t(t)-\phi_{r}^h(t) = \left( \omega^t -\omega^h \right)t,
\end{equation}
which will also be incorporated into the gait equation.

\begin{figure}[tb!]
    \centering
    \includegraphics[width=1\linewidth]{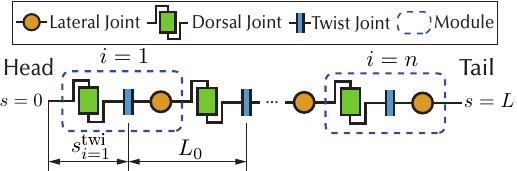}
    \caption{Illustration of the discrete mechanism of \abbrTSnake.
    }
    \label{fig:mechanism}
\end{figure}

\subsection{Discrete Mechanism of \abbrTSnake}\label{subsec:mechanism}

As shown in Fig.~\ref{fig:mechanism}, the \abbrTSnake mechanism consists of $n$ identical modules, each containing three 1-DoF revolute joints arranged in a \textit{dorsal-twist-lateral} sequence.
The dorsal and lateral joint axes pass through the cross-sectional center and span the normal plane of the backbone.
They are mutually orthogonal in the straight configuration.
The twist joint axis is locally aligned with the backbone tangent and is therefore orthogonal to both bending axes.

Each module has length $L_0$. 
The complete mechanism has $3n$ DoFs and a total length of $L=nL_0$.
These three joint axes provide a discrete realization of the dorsal reference curve: the lateral and dorsal joint rotations represent the accumulated curvatures $\kappa_a$ and $\kappa_b$, respectively, while the twist joint rotation represents the accumulated torsion $\tau$.

\subsection{Gait Equations}\label{subsec:gait_eqn}

To derive the gait equation of the S-rolling, we apply the integration method from \cite{yamada2010approximations} on the \abbrTSnake. 
As we discussed in Sec.~\ref{subsec:backbone_curves}, when the head and tail segments undergo different rolling motions, a discontinuous twist angle $\alpha(t)$ arises at the midpoint of the \abbrTSnake body, i.e, at $s=L/2$.
This relative twist must be physically realized by a twist joint of \abbrTSnake.
We therefore consider a \abbrTSnake configuration with an odd number of modules, $n=2k+1,k\in\mathbb{Z}^+ $.
Since the twist joint is located at the center of each module (Fig.~\ref{fig:mechanism}), the body midpoint $s=L/2$ of a \abbrTSnake with an odd number of modules coincides with the twist joint of the $(k+1)$-th module.
This central twist joint is thus used to realize the twist joint between the head and tail segments.

Let $\theta_i$ and $s_i$ be the joint angle and position measured by arc-length $s$ of the $i$-th joint of the same type (lateral, dorsal, or twist) along the \abbrTSnake.
Recall that $L_0$ is the length of each module of \abbrTSnake (Sec.~\ref{subsec:mechanism}).
Hence, $L_0$ is also the distance between two adjacent joints of the same type.
Then, by integrating the shape functions (i.e., curvatures)  along the curve between two adjacent dorsal or lateral joints, the corresponding joint angles can be obtained, which can then be used to control the \abbrTSnake.
Here, we substitute the shape functions derived in Eqs.~\eqref{eqn:new_shape_functions} into the integration approach.
For the dorsal joints:
\begin{subequations}\label{eqn:gait_dorsal}
    \begin{align}
        \theta^\text{dor}_i(t) &= 
        \begin{cases}
        \displaystyle
            \int_{s_i^\text{dor}-L_0/2}^{s_i^\text{dor}+L_0/2} \kappa_b^h(t) \, {\rm d}s, & 0 \leq s_i^\text{dor} \leq \frac{L}{2}, \\[2mm]
            \displaystyle
            \int_{s_i^\text{dor}-L_0/2}^{s_i^\text{dor}+L_0/2} \kappa_b^t(t) \,  {\rm d}s, & \frac{L}{2} < s_i^\text{dor} \leq L,
        \end{cases}\\
        & =
        \begin{cases}
            \kappa_0 L_0 \cos{\left( \omega^h t \right)}, & 1 \leq i \leq k+1, \\[1mm]
            -\kappa_0 L_0 \cos{\left( \omega^t t \right)}, & k+1 < i \leq 2k+1.
        \end{cases}\label{eqn:angle_dorsal}
    \end{align}
\end{subequations}
Similarly, for the lateral joints:
\begin{subequations}\label{eqn:gait_lateral}
    \begin{align}
        \theta^\text{lat}_i(t) &= 
        \begin{cases}
        \displaystyle
            \int_{s_i^\text{lat}-L_0/2}^{s_i^\text{lat}+L_0/2} \kappa_a^h(t) \, {\rm d}s, & 0 \leq s_i^\text{lat} \leq \frac{L}{2}, \\[2mm]
            \displaystyle
            \int_{s_i^\text{lat}-L_0/2}^{s_i^\text{lat}+L_0/2} \kappa_a^t(t) \, {\rm d}s, & \frac{L}{2} < s_i^\text{lat} \leq L,
        \end{cases}\\
        & =
        \begin{cases}
            \kappa_0 L_0 \sin{\left( \omega^h t \right)}, & 1 \leq i < k+1, \\[1mm]
            -\kappa_0 L_0 \sin{\left( \omega^t t \right)}, & k+1 \leq i \leq 2k+1.
        \end{cases}\label{eqn:angle_lateral}
    \end{align}
\end{subequations}
For the twist joints, as we mentioned before, the $(k+1)$-th twist joint should realize the twist angle $\alpha(t)$, while the others remain zero:
\begin{equation}
    \label{eqn:angle_twist}
    \theta^\text{twi}_i(t)=
    \begin{cases}
        \left( \omega^t -\omega^h \right)t, & i=k+1,\\
        0, & i \neq k+1.
    \end{cases}
\end{equation}
Equations~\eqref{eqn:angle_dorsal}, \eqref{eqn:angle_lateral}, and \eqref{eqn:angle_twist} constitute the gait equation for S-rolling, which governs the joint angles of the three joint types in the \abbrTSnake.

\begin{figure}[tb!]
    \centering
    \includegraphics[width=0.6\linewidth]{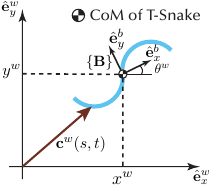}
    \caption{Illustration of the kinematic model of S-rolling on the \abbrTSnake.
    }
    \label{fig:kinematic}
\end{figure}

\subsection{Kinematic Modeling}\label{subsec:kinematic_model}

The kinematic model of S-rolling is derived under a quasi-static approximation~\cite{hu2009mechanics}.
The key assumptions adopted in deriving the model are summarized below.

\begin{assumption} [Uniform density] \label{asp:density}
    The mass per unit arc-length of \abbrTSnake is assumed to be constant. 
    Therefore, its center of mass (CoM) coincides with the geometric centroid of the backbone curve.
\end{assumption}

To derive the resultant velocity of the \abbrTSnake,  we make the following assumption to determine the velocity direction and magnitude at each point along its backbone curve.

\begin{assumption} [Segment rolling speed] \label{asp:roll_speed}
Within each segment, the rolling speed is spatially invariant with respect to arc-length $s$, and its direction is defined by the normal vector $\hat{\mathbf e}_a^r$ of the bellows frame $\mathbf D^r$.
\end{assumption}

The \abbrTSnake moves in the $XoY$ plane of the world frame $\{\mathbf{W} \}$.
At $t=0$, its CoM is located at the origin of $\{\mathbf{W} \}$ (see Sec.~\ref{subsec:backbone_curves}).
We define a 2D body frame $\{\mathbf{B} \}=[\hat{\mathbf e}_x^b,\hat{\mathbf e}_y^b]^\top \in \mathbb{R}^{2\times 2}$, whose origin is attached to the CoM of \abbrTSnake (Fig.~\ref{fig:kinematic}).
Initially, $\hat{\mathbf e}_x^b$ and $\hat{\mathbf e}_y^b$ are aligned with $\hat{\mathbf e}_x^w$ and $\hat{\mathbf e}_y^w$, respectively.
The pose of the body frame with respect to the world frame also represents the CoM pose of \abbrTSnake, which is defined as $\mathbf{q}(t)=[x^w,y^w,\theta^w]^\top$, where $\theta^w$ denotes the angle measured from the positive $x$-axis of the world frame to the positive $x$-axis of the body frame (Fig.~\ref{fig:kinematic}).

\begin{proposition} [Kinematic model of S-rolling] \label{pro:kinematic model}
The kinematic model of S-rolling expressed in the world frame is
\begin{equation}
    \label{eqn:kinematic_model}
    \dot{\mathbf{q}} =
    \begin{bmatrix}
        \dot{x^w} \\ 
        \dot{y^w} \\
        \dot{\theta^w} 
    \end{bmatrix} 
     = 
    \begin{bmatrix}
        -C_1 \sin{\theta^w} & -C_1 \sin{\theta^w} \\
        C_1 \cos{\theta^w} & C_1 \cos{\theta^w} \\
        -C_2 & C_2
    \end{bmatrix}
    \begin{bmatrix}
        \omega^h \\
        \omega^t
    \end{bmatrix} ,
\end{equation}
where $C_1$ and $C_2$ are coefficients defined below.

\end{proposition}

The derivation of this model proceeds as follows.
After the planar motion, the backbone curve of \abbrTSnake expressed in the world frame is given by:
\begin{equation}
    \label{eqn:curve_in_W}
    \mathbf{c}^w(s,t)= [x^w,y^w]
    + [\mathbf{c}^r(s)]_{xy} \mathbf{R}^\top(\theta^w) \in \mathbb{R}^{1\times2},
\end{equation}
where $[\cdot]_{xy}$ denotes the operator that extracts the $x$- and $y$-components of a vector, and $\mathbf R$ is the 2D rotation matrix from the body frame to the world frame:
\begin{equation}\label{eqn:rotation_mat}
    \mathbf R(\theta^w)
    =
    \begin{bmatrix}
    \cos\theta^w&-\sin\theta^w\\
    \sin\theta^w&\cos\theta^w
\end{bmatrix}.
\end{equation}

We denote the cross-sectional radius of \abbrTSnake as $r_c=A_{cd}/2$.
Based on Assumption~\ref{asp:roll_speed}, the velocity field $\mathbf{u}(s,t) \in \mathbb{R}^{1 \times 3}$ of \abbrTSnake expressed in the body frame is given by:
\begin{equation}
    \label{eqn:vel_field}
    \mathbf{u}(s,t)=\begin{cases}
    \displaystyle
        r_c \omega^h(t) \hat{\mathbf e}_a^r(s), &0 \leq s \leq \frac{L}{2} \\[1mm]
    \displaystyle
         r_c \omega^t(t) \hat{\mathbf e}_a^r(s), &\frac{L}{2} < s \leq L
    \end{cases}
\end{equation}
where $\hat{\mathbf e}_a^r$ is the normal vector of the frame $\mathbf D^r$ in Eqs.~\eqref{eqn:psi_to_vectors}.
The CoM velocity expressed in the body frame, denoted by $\mathbf{v}^b$, is given by:
\begin{equation}
    \label{eqn:vel_body_CoM}
    \mathbf{v}^b=\frac{1}{L}\int_0^L [\mathbf{u}]^\top_{xy} \, {\rm d}s = \begin{bmatrix}
        0 \\
        C_1(\omega^h+\omega^t)
    \end{bmatrix}.
\end{equation}
where we denote the coefficient $C_1=\frac{2 r_c \sin{\left(\kappa_0 L / 4 \right)}}{\kappa_0 L}$ for simplicity. 
Since the planar angular velocity is directed along the common $z$-axis of the body and world frames, its scalar value is identical in both frames.
Therefore, the angular velocity of \abbrTSnake is given by:
\begin{equation}
    \label{eqn:angular_vel}
    \dot{\theta}^w=\frac{\int_0^L [\mathbf{c}^r \times \mathbf{u} ]_z \, {\rm d}s}{\int_0^L\|\mathbf{c}^r\|^2 \, {\rm d}s} 
    = C_2(\omega^t-\omega^h),
\end{equation}
where $[\cdot]_z$ denotes the $z$-component of a vector and therefore yields a scalar, $\times$ denotes the vector cross product, and the denominator represents the arc-length polar second moment of the \abbrTSnake backbone about its CoM.
Similarly, we denote the coefficient $C_2=\frac{r_c \kappa_0 \sin^2{\left(\kappa_0 L / 4 \right)}}{\kappa_0 L- 2\sin{\left(\kappa_0 L / 2 \right)}}$ for simplicity in Eq.~\eqref{eqn:angular_vel}.
The final kinematic model of S-rolling under the world frame can be derived as follows:
\begin{equation}
    \label{eqn:kinematic_model_derive}  
    \dot{\mathbf{q}} =
    \begin{bmatrix}
        \dot{x^w} \\ 
        \dot{y^w} \\
        \dot{\theta^w} 
    \end{bmatrix} = 
    \begin{bmatrix}
        \mathbf{R}(\theta^w) \mathbf{v}^b \\
        \dot{\theta^w} 
    \end{bmatrix}. 
\end{equation}
The fully expanded form of Eq.~\eqref{eqn:kinematic_model_derive} is given in Eq.~\eqref{eqn:kinematic_model}.

As indicated by Eq.~\eqref{eqn:kinematic_model}, different motion modes can be generated by independently controlling the rolling speeds of the head and tail segments.
When $\omega^h=\omega^t$, the angular velocity vanishes and \abbrTSnake moves along a straight path.
When $\omega^h\neq\omega^t$, the difference between the two rolling speeds produces a turning motion. 
Hardware demonstrations of these motion modes are presented in Sec.~\ref{subsec:modes}.


%% file: experiments.tex
This section shows the results of real-robot experiments of S-rolling on the \abbrTSnake.
We first illustrate various motion modes generated by different combinations of rolling speeds (Sec.~\ref{subsec:modes}).
Next, we drive \abbrTSnake along a trajectory on a more challenging test track to demonstrate its maneuverability (Sec.~\ref{subsec:maneuverability}).
Finally, we consider two real-world LCEs, design corresponding tasks for \abbrTSnake, and accomplish them using the S-rolling gait.

It is worth noting that the twist joints of the \abbrTSnake hardware impose no torsional angle limits.\footnote{Our physical \abbrTSnake implements the mechanism in Fig.~\ref{fig:mechanism} as described in Sec.~\ref{subsec:mechanism}, and the design of this robot is neither the focus nor claimed as a contribution of this paper.}
Therefore, $\theta_i^{\mathrm{twi}}$ in Eq.~\eqref{eqn:angle_twist} never reaches any upper or lower bound, and the head and tail segments can twist relative to each other without restriction.
In each group of experiments, \abbrTSnake is composed of a different number $n$ of modules, and different rolling speeds $\omega^h$ and $\omega^t$ are employed to generate distinct motion modes.
These parameters are specified in the respective subsections.
The common parameters are as follows: the module length $L_0$ is $0.2055$ m, the host computer updates the control commands for T-Snake at $50$ Hz, and the angular amplitude $\kappa_0 L_0$ in Eqs.~\eqref{eqn:angle_dorsal} and \eqref{eqn:angle_lateral} is $0.5$ rad.

\begin{figure}[tb!]
    \centering
    \includegraphics[width=1\linewidth]{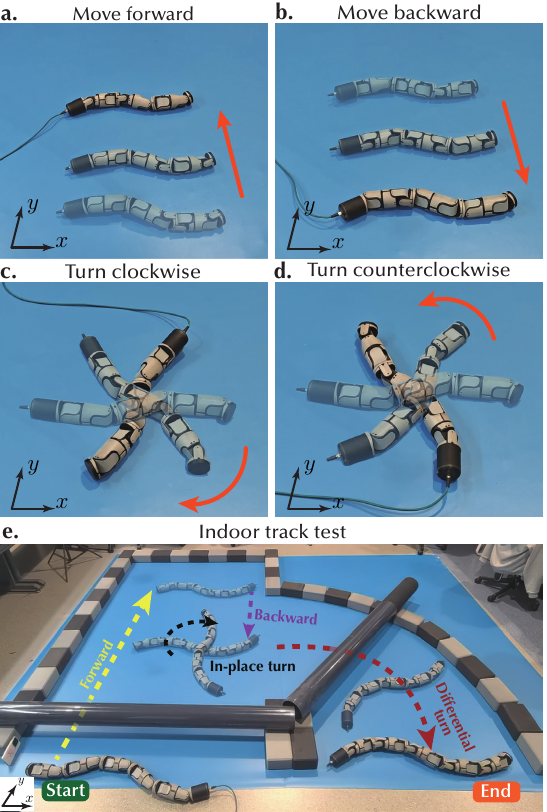}
    \caption{S-rolling locomotion tests under laboratory conditions. (a)-(d) Different motion modes of S-rolling on the \abbrTSnake hardware. (e) Trajectory of the \abbrTSnake traversing the test track.}
    \label{fig:indoor_tests}
\end{figure}

\subsection{Different Motion Modes}\label{subsec:modes}

This experiment validates the translating and turning capabilities of the S-rolling gait.
The \abbrTSnake hardware consists of three modules, with a total of $3\times3=9$ DoFs.
Experiments are conducted indoors on a flat, hard surface.

\subsubsection{Translating}

According to the kinematic model of S-rolling (Eq.~\eqref{eqn:kinematic_model}), the angular velocity of \abbrTSnake, $\dot{\theta^w}$, vanishes when the head and tail rolling speeds are equal.
We set $\omega^h=\omega^t=0.5$ rad/s and $\omega^h=\omega^t=-0.5$ rad/s to generate forward and backward motions.
The resulting motions are shown in Fig.~\ref{fig:indoor_tests}(a) and (b). 
Ten velocity measurements are performed for each motion. 
The kinematic model (Eq.~\eqref{eqn:kinematic_model}) predicts a linear velocity of $1.75$ cm/s, whereas the measured average forward and backward velocities are $(1.86 \pm 0.20)$ cm/s and $(1.97 \pm 0.19)$ cm/s, corresponding to relative errors of $5.7\%$ and $10.9\%$, respectively.

\subsubsection{In-place Turning}

According to Eq.~\eqref{eqn:kinematic_model}, when $\omega^h$ and $\omega^t$ are opposite, the linear velocity of \abbrTSnake becomes zero, causing the robot to rotate in place about its CoM.
We set $\omega^h=-\omega^t=0.5$ rad/s to generate clockwise turning motion, and $\omega^h=-\omega^t=-0.5$ rad/s to generate counterclockwise turning motion.
The resulting motions are shown in Fig.~\ref{fig:indoor_tests}(c) and (d). 
The kinematic model predicts an angular velocity of $4.19$ $^\circ$/s, whereas the measured average angular velocities are $(8.03 \pm 0.72)$ $^\circ$/s and $(7.54 \pm 0.50)$ $^\circ$/s for clockwise and counterclockwise turning, corresponding to relative errors of $47.9\%$ and $44.5\%$, respectively.
We attribute this large discrepancy between the model prediction and measurements to the slip between the \abbrTSnake body and the ground during in-place turning, which does not fully conform to the quasi-static assumption adopted in our model.
The further analysis of this effect will be left for future work.

\begin{figure*}[tb!]
    \centering
    \includegraphics[width=1\linewidth]{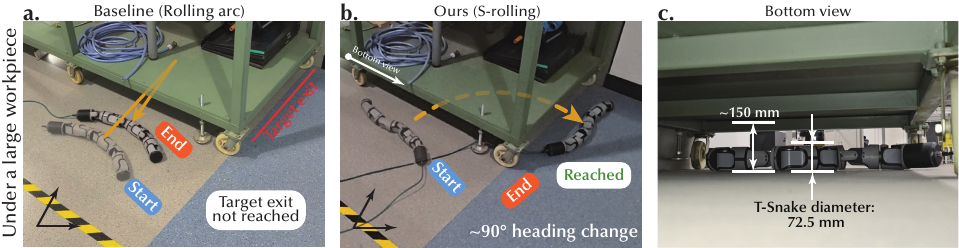}
    \caption{Illustration of the \abbrTSnake tests under a large workpiece. (a) The rolling arc gait cannot reach the target exit, while (b) the proposed S-rolling gait completes the task successfully. (c) The bottom view of the \abbrTSnake traversing under the workpiece.}
    \label{fig:workpiece}
\end{figure*}

\subsection{Maneuverability in Low-Clearance Environments}\label{subsec:maneuverability}

An indoor test track is constructed on a flat and hard floor to evaluate the maneuverability of the S-rolling gait in LCEs.
The track measures $3~\mathrm{m} \times 4~\mathrm{m} $ and includes two LCEs formed using pipes, each with a clearance height of $10$ cm.
At the start of the tests, we use the sidewinding gait to move the \abbrTSnake onto the track, then switch to S-rolling.
The motion modes of S-rolling are controlled by a joystick, allowing a human operator to teleoperate the \abbrTSnake during the test.
The \abbrTSnake hardware in this experiment consists of five modules, yielding a total of $3 \times 5 = 15$ DoFs.
\footnote{Using 5 modules (as opposed to 3) helps the snake robot to achieve sidewinding reliably to quickly reach the entrance of the LCE in this test.
Note that, this test verifies that our new S-rolling gaits provide the snake robot with new additional capability in LCEs, without losing or disabling existing gaits (e.g. sidewinding) of a snake robot.}

During this experiment, the \abbrTSnake first moves forward in a straight line and then reverses.
It subsequently executes an in-place turn of approximately $90^\circ$, followed by differential steering through a curved track (Fig.~\ref{fig:indoor_tests}(e)).
The complete motion sequence is provided in the supplementary video.

\subsection{Real-world Scenario Experiments}\label{subsec:real-world}

We design tasks in real-world LCEs to show the practical utility of S-rolling.
We compare S-rolling with an existing rolling gait, rolling arc~\cite{zhen2015modelinga}, by evaluating their task completion performance across two tasks.
The \abbrTSnake hardware used in these experiments comprises three modules, and a human operator teleoperates the robot to complete the tasks.

\subsubsection{Under a Large Workpiece}

When an LCE serves as the only passage connecting two spaces, the gait must maneuver flexibly in the LCE.
We design a task in which the \abbrTSnake enters from one side of a large workpiece with a clearance height of $15$ cm and exits from an orthogonal side.
Rolling arc can move only along a straight path and cannot steer (Fig.~\ref{fig:workpiece}(a)), whereas S-rolling can steer flexibly (Fig.~\ref{fig:workpiece}(b)).

\subsubsection{Under a Car Chassis}

Many LCEs require inspection and nondestructive testing.
We design a task in which the \abbrTSnake enters beneath a car chassis with a height of $13$ cm, performs an in-place $180^\circ$ turn to simulate inspection data acquisition, and then exits (Fig.~\ref{fig:head_figure}(b)). 
The baseline gait cannot change the heading of the \abbrTSnake (Fig.~\ref{fig:head_figure}(c)), whereas S-rolling completes the task successfully (Fig.~\ref{fig:head_figure}(d)).

%% file: conclude.tex
This paper presented S-rolling, a steerable rolling gait for twistable snake robots operating in low-clearance environments. 
We design the backbone curve of the gait, and derive a joint-space gait equation for the \abbrTSnake in which a central twist joint realizes the relative twisting. 
A quasi-static kinematic model is derived, relating the two segment rolling speeds to the robot's linear and angular velocities.
Hardware experiments demonstrate different motion modes, traversal of a $10$ cm-clearance test track, and real-world tasks beneath a workpiece and a car chassis, confirming that S-rolling combines the low swept height of planar rolling with the steering capability that existing rolling gaits lack.

Future work will extend S-rolling from teleoperation to closed-loop autonomous navigation in low-clearance environments. 
We also plan to integrate onboard sensing for inspection and nondestructive-testing tasks.